\documentclass[a4, 8pt]{nature}
\usepackage{geometry}
\usepackage{graphicx}
\usepackage{color}
\usepackage{xurl}
\usepackage{hyperref}
\usepackage{amsfonts}
\usepackage{amsmath}
\usepackage{amssymb}
\usepackage{algorithm}
\usepackage{algorithmic}
\usepackage{caption}
\usepackage{graphics}

\usepackage{upgreek}
\usepackage{multirow}
\usepackage{diagbox}
\usepackage{xspace}
\usepackage{fancyhdr}
\usepackage{makecell}

\DeclareCaptionLabelSeparator{naturebar}{\textbf{\,\textbar\,}}

\definecolor{darkred}{rgb}{0.8, 0.2, 0.2}
\definecolor{darkgreen}{rgb}{0, 0.8, 0}
\definecolor{darkpurple}{rgb}{1.0, 0, 1.0}
\definecolor{darkblue}{rgb}{0, 0, 1.0}
\definecolor{orange}{rgb}{1, 0.5, 0}

\renewcommand{\thefootnote}{(\roman{footnote})} 

\newcommand{\beginsupplement}{%
        \setcounter{table}{0}
        \renewcommand{\thetable}{\arabic{table}}%
        \setcounter{figure}{0}
        \renewcommand{\thefigure}{\arabic{figure}}%
     }

\newcommand{\R}		{\mathbb{R}}	
\newcommand{\DensityMap}       {\rho}      
\newcommand{\SimpComplex}       {\mathsf{K}} 

\newcommand{\PersistenceThreshold} {\tau}

\newcommand{\TreeRoot}{v_{0}}
\newcommand{\SimpThreshold} {\tau}
\newcommand{\SimpDensityScore} {dscore}
\newcommand{\SimpVecScore} {vscore}
\newcommand{\SimpWeight} {\alpha}
\newcommand{\SimpWeightScore} {s}
\newcommand{\SimpFinalScore} {\overline{s}}
\newcommand{\SimplifiedTree}{T_s}
\newcommand{\WeightSummaryDistCap}{\beta_w}
\newcommand{\ThicknessDistCap} {\beta_t}
\newcommand{\SimpDensityDistCap} {\beta_s}

\newcommand{\leafburner}{\texttt{LeafBurner}\xspace}
\newcommand{\rootgrower}{\texttt{RootGrower}\xspace}
\newcommand{\DiscreteMorse}{\texttt{DiMorSC}\xspace}

\newcommand{\myDM}      {{DM-Skeleton}\xspace} 

\title{\Huge Detection and skeletonization of single neurons and tracer injections using topological methods}
\date{}

\author{\Large
    \textbf{Dingkang Wang$^1$, Lucas Magee$^1$, Bing-Xing Huo$^2$, Samik Banerjee$^2$, Xu Li$^2$, Jaikishan Jayakumar$^3$, Meng Kuan Lin$^2$,  Keerthi Ram$^3$, Suyi Wang$^1$, Yusu Wang$^1$, Partha P. Mitra$^2$}}

\newcommand\blfootnote[1]{%
  \begingroup
  \renewcommand\thefootnote{}\footnote{#1}%
  \addtocounter{footnote}{-1}%
  \endgroup
}

\begin{document}

\captionsetup[figure]{labelfont={bf},labelformat={default},labelsep=naturebar,name={Fig.}}

\twocolumn[{%
\begin{@twocolumnfalse}
\maketitle
\vspace{5mm}

\begin{abstract}
Neuroscientific data analysis has traditionally relied on linear algebra and stochastic process theory. However, the tree-like shapes of neurons cannot be described easily as points in a vector space (the subtraction of two neuronal shapes is not a meaningful operation), and methods from computational topology are better suited to their analysis. Here we introduce methods from Discrete Morse (DM) Theory to extract the tree-skeletons of individual neurons from volumetric brain image data, and to summarize collections of neurons labelled by tracer injections. Since individual neurons are topologically trees, it is sensible to summarize the collection of neurons using a consensus tree-shape that provides a richer information summary than the traditional regional ‘connectivity matrix’ approach. The conceptually elegant DM approach lacks hand-tuned parameters and captures global properties of the data as opposed to previous approaches which are inherently local. For individual skeletonization of sparsely labelled neurons we obtain substantial performance gains over state-of-the-art non-topological methods (over 10\% improvements in precision and faster proofreading). The consensus-tree summary of tracer injections incorporates the regional connectivity matrix information, but in addition captures the collective collateral branching patterns of the set of neurons connected to the injection site, and provides a bridge between single-neuron morphology and tracer-injection data.\\
\end{abstract}
\vspace{5mm}
\end{@twocolumnfalse}
}]

\section*{Summary}
\blfootnote{
\begin{affiliations}
 \item Computer Science and Engineering Department, The Ohio State University, Columbus, OH, USA 43210.
 \item Cold Spring Harbor Laboratory, NY, USA 11724.
 \item Indian Institute of Technology, Chennai, Tamil Nadu, India 600036.
\end{affiliations}
}
\indent Neuroscientific data analysis has traditionally involved methods for statistical signal and image processing, drawing on linear algebra and stochastic process theory. However, digitized neuroanatomical datasets containing labelled neurons, either individually or in groups labelled by tracer injections, do not fit into this classical framework. The tree-like shapes of neurons cannot be adequately described as points in a vector space (e.g. the subtraction of two neuronal shapes is not a meaningful operation). There is therefore a need for new approaches, which has become more urgent given the growth in whole-brain datasets with sparsely labelled neurons or tracer injections.\\
\indent Methods from computational topology and geometry are naturally suited to the analysis of neuronal shapes. In this paper we introduce methods from Discrete Morse Theory to extract tree-skeletons of individual neurons from volumetric brain image data, and to summarize collections of neurons labelled by anterograde tracer injections. Since individual neurons are topologically trees, it is sensible to summarize the collection of neurons using a consensus tree-shape. This consensus tree provides a richer information summary than the regional or voxel-based ”connectivity matrix” approach that has previously been used in the literature. \\
\indent The algorithmic procedure for single neuron skeletonization includes an initial neurite-detection step to extract a density field from the raw volumetric image data, followed by Discrete Morse theoretic skeleton extraction from the density field using the 1-(un)stable manifold of the density field. We apply the method to high-resolution 3D volumes of sparsely-labelled neurons to quantitatively extract the single neuron topology, and achieve better performance than state-of-the-art algorithms using non-topological methods (over 10\% improvement in precision, and significant speedups at the proofreading stage). \\
\indent In image volumes of  brains with neuronal tracer injections, a summarized {\it consensus tree} of the collective neuronal projection patterns is extracted to characterize brain-wide neuronal connectivity. The procedure for building the tree skeleton assigns the weight associated with detected tracer projections locally to the nearest point on the skeleton. This ensures that the resulting weighted skeleton also provides a summary of the tracer projection data: integrating the weighted skeleton over a brain region, helps recover the regional "connectivity matrix". Thus the consensus-tree summary of tracer injections incorporates the traditional regional connectivity matrix information, but in addition captures the collective {\it collateral branching patterns} of the set of neurons connected to the injection site. We propose that this summary can provide a future means of characterizing tracer injection data, and also provide a bridge to a growing body of single-neuron morphology data. \\
\indent We find that the DM method is able to trace a tree branch through regions of low intensity that pose challenges to baseline methods. This is a particular strength of the DM approach as it utilizes the global topological structure present in the data, whereas relevant literature methods are inherently spatially local. Additionally the DM approach is theoretically well principled and conceptually clean without multiple ad-hoc hand-engineered steps. There is a significant computational overhead to the topological approaches but we are able to mitigate the speed issues using parallelized implementations of the core algorithms.  

\section*{Background}
\indent {\bf Topological Data Analysis and Discrete Morse Theory:}
Within the last decade or so, topological data analysis (TDA) has grown in influence. Several new TDA methodologies have been proposed for analyzing complex high-dimensional datasets. These approaches use a variety of topological concepts to characterize essential structure behind the data and algorithmic implementations have been developed \cite{EH10,Carl09,CM17,LS13,Tie18}. Some of the ideas (in particular persistent homology \cite{Edelsbrunner2002}) have been applied to multiple application domains \cite{Buchet2018,dario,platt,Lamar,LBD17}, including neuroscience \cite{li2017,CG19,KDS2018}. \\
\indent The area of TDA that relates to the present work is the persistence-guided Discrete Morse theory based framework for reconstructing hidden graphs from observed data. Discrete Morse theory has been utilized to capture hidden structure in 2D or 3D volumetric data \cite{DRS15,GDN07,RWS11}. Extraction of hidden graphs was formulated in \cite{2011MNRAS}, and the framework was simplified and theoretical guarantees provided in \cite{DWW18}. 
Morse theory exploits the global (instead of local) topology of a scalar field. Thus the hidden graph skeleton of the scalar field can be extracted reliably through regions of weak signals. Classical Morse theory applies to continuous functions on manifolds. Discrete Morse theory \cite{Forman_DM_1998} provides a discretized computational framework suitable for algorithmic implementation on digitized data. Persistent homology is used to separate signal from noise and remove potentially noise-related structure from the graph. Such a framework has been applied previously to reconstructing hidden road networks from noisy GPS trajectories and satellite images \cite{WWL15,DWW19}. Here we adapt and extend this approach to develop a computational methodology suitable for computational neuroanatomy and to address neuroscientific problems.\\
\indent In this manuscript we introduce a data analysis framework entitled DM-skeleton that uses TDA and the Discrete Morse approach to skeletonize individual neurons as well as groups of neurons labelled by tracer injections. For single neuron reconstruction, we show that DM-skeleton significantly outperforms the best available automated approaches\cite{APP2_Hang2013, Hang_GTree2018} (over $10\%$ improvements in F1 scores) and reduces proofreading times. For tracer injection skeletonization, DM-skeleton provides a conceptually new route to the analysis of mesoscale projection data, and shows robust performance. \\
\indent {\bf Single Neuron Reconstruction:} Many methods have been previously proposed for single-neuron reconstruction from high resolution image stacks \cite{Al-Kofahi:2002, Bas2011, Basu2014, Boykov2001, Choromanska2012, Chothani2011, activelearning2014, Lee2008, lee2012, Myatt2012, oh2014mesoscale, sironi2015, SCHMITT20041283, sui2014, Srinivasan2007, turetken2013, Turetken2011, VASILKOSKI2009197, Yang2013, ZHANG2007149, Zhao2011, Zhou1999, Zhou2015, Zhou2016, APP2_Hang2013, smarttracing, Quan_NeuroGPSTree_2016, Hang_GTree2018} (see also surveys and books \cite{Acciai2016, DONOHUE201194, neuronsurvey, arenkiel2014neural} for more detailed discussions.) Most of the existing methods sequentially expand a tree from a collection of seed points that are often selected based on density information. 
A popular line of sequential tracing algorithms use shortest-path  based  approaches, such as APP2 \cite{APP2_Hang2013} and SmartTracing\cite{smarttracing}. 
More recently, methods using a principal-curve to sequentially include more nodes in the reconstruction (e.g., NeuroGPS-Tree \cite{Quan_NeuroGPSTree_2016} and GTree \cite{Hang_GTree2018}) have shown high performance on single neuron reconstruction in mouse brain datasets. Our proposed DM-skeleton approach shows robust performance gains over APP2 and GTree, and shortens proofreading time. Importantly, it also opens up a conceptually new direction in single neuron tracing by using global topological information in a manner that we feel will generalize beyond this specific work.\\
\indent {\bf Tracer Injection Skeletonization:} Brain-wide tracer injection datasets are another area where we apply the DM skeletonization approach. In tracer injection data sets, thousands of neurons are collectively labelled and do not have the topological simplicity of individual trees. Conventionally, brain-wide connectivity information is summarized in the form of regional connectivity matrices \cite{ConnectivityMatrix_chapter3_2016}.
Such a representation ignores the fundamental tree-like morphology of neurons and contains no information relating to collateral branching patterns. In this manuscript we introduce a new approach to the analysis of tracer injection data, by {\it skeletonizing} the tracer injections using a {\it summary tree}. To the best of our knowledge, this is a new approach to conceptualizing tracer-injection data, and could provide a biologically better-grounded approach to the study of mesoscale connectivity mapping using tracer injections. One important advantage of the tree-skeletonization approach is that it connects more directly with single-neuron reconstructions. \\
\indent It is possible to utilize methods such as APP2 or GTree to perform tracer-injection skeletonization (and we use these as baseline comparison techniques). However, these methods do not automatically interpolate through regions of weak label, a problem that is exacerbated for tracer-injection data. In contrast, the DM based approach utilizes global information and is more robust.\\
\begin{figure*}
    \centering
	\includegraphics[width = 0.8\linewidth]{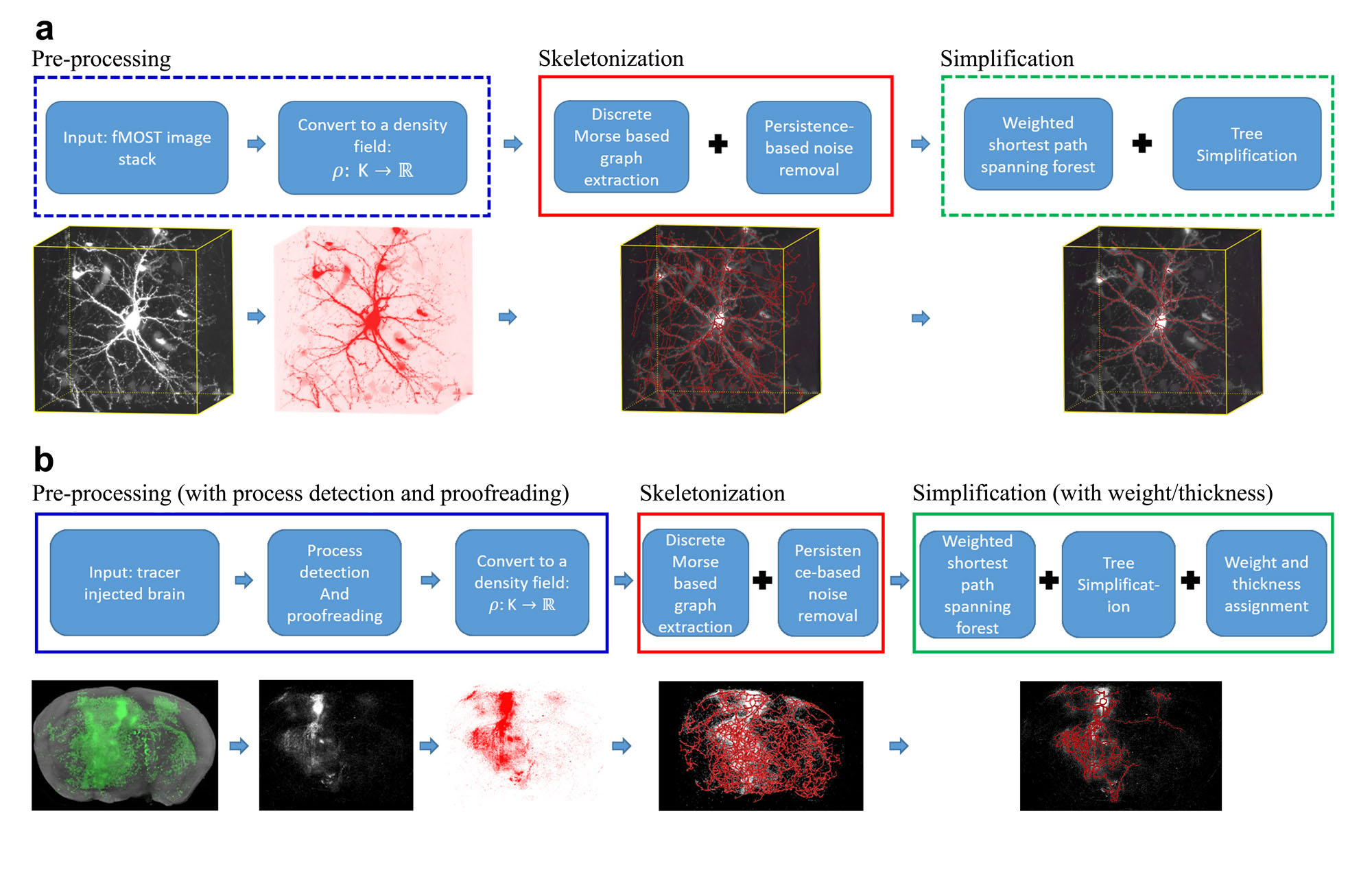}
	\caption{Workflow and sample outputs for the \myDM{} pipeline.  \textbf{a,} Workflow for fMOST datasets (from sparse neuronal label to individual skeleton extraction) and sample outputs for each step.
 	\textbf{b,} Workflow for tracer injection datasets (from dense neuronal label to injection summarization) and sample outputs for each step.}
	\label{fig:1_workflows}
\end{figure*}
\indent The DM pipeline for summarizing multiple-neuron tracer injection datasets has multiple stages. First, the raw image stack is preprocessed to detect or highlight neuronal processes \cite{Samik_DM++}. Then a variant of the Discrete-Morse module \cite{WWL15,DWW18} is employed to produce a graph skeleton containing all potential neuronal trajectories. This graph skeleton may contain false positives, so the next stage of our algorithm performs a persistent homology based simplification step which removes superfluous branches in low-density regions and branches that misalign with estimated ``flow vectors''. This simplified graph is skeletonized into a minimum spanning tree, which is further simplified using a persistence threshold. During all simplification steps, weights assigned to points on the tree (summarizing the neighboring projection density) is re-assigned so as to preserve the total weight. The resulting weighted tree-summary provides a new way of characterizing tracer injection data, simultaneously capturing regional connectivity information as well as collateral branching patterns.  
\section*{Results} \label{sec:Results}
\begin{figure*}
	\centering
    \includegraphics[width = 0.8\linewidth]{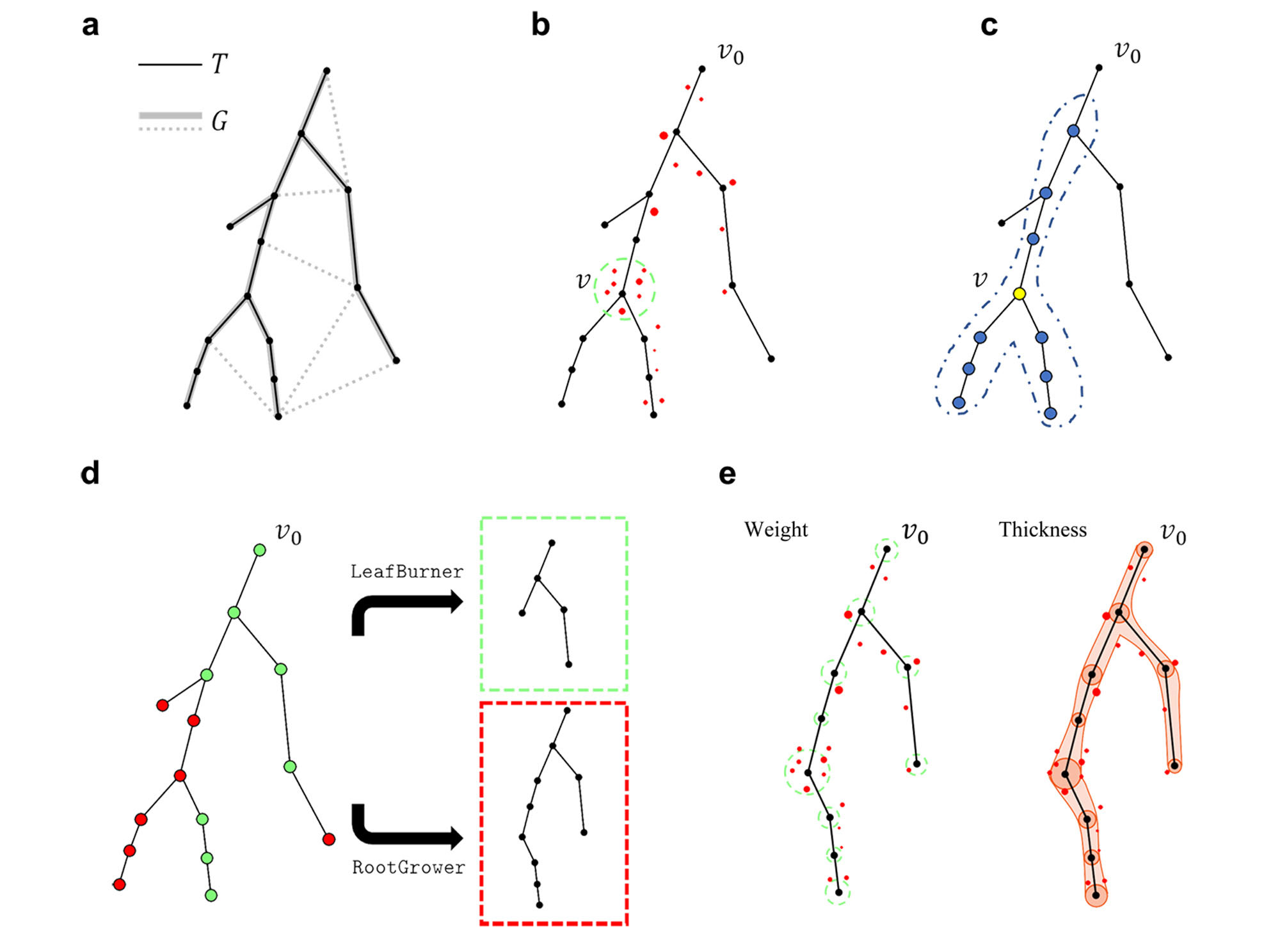}
	\caption{Summarization and simplification algorithms. \textbf{a,} An example of spanning forest algorithm. Only one of the trees is shown for simplicity. Assume $G$ is the graph output from Discrete Morse, and $T$ is its weighted shortest path spanning tree. The edges are weighted by the average intensity of both endpoints - larger weights mean smaller distances. Dotted lines represent those edges that are removed from $G$. \textbf{b,} Score assignment for tree nodes. Each voxel is associated with its nearest neighbor in the set of tree nodes. The score of a tree node $v$ is the sum of the density of voxels associated with $v$. An upper bound $\SimpDensityDistCap$ is set to avoid distant associations. \textbf{c,} Score smoothing: To reduce noise, we smooth the scores of each tree node $v$ by averaging those of its neighbors within $k$-hops (i.e., connected via at most $k$ edges to $v$). We restrict only to neighbors that are ancestors or descendants of $v$. An example with $k=3$ is shown in the figure, where blue nodes are neighbors of node $v$. \textbf{d,} Simplification process: We utilize two strategies. Strategy 1 (\texttt{LeafBurner}) starts from the leaves and iteratively removes tree nodes with scores $< \SimpThreshold$. Strategy 2 (\texttt{RootGrower}) grows the tree from the root $\TreeRoot$ by keeping tree nodes with scores $> \SimpThreshold$. In the tree on the left, green nodes (red nodes) correspond to nodes with score higher (lower) than the threshold $\tau$. Deploying different strategies will result in different outputs. \textbf{e,} To achieve a weight-preserving summarization, each tree node has a weight (represented by the size of the dotted green circle) equal to the sum of the weights of associated voxels. We also assign a thickness value to each tree node (represented by the radius of the orange circle) for better visualization. This thickness value is proportional to the square root of the number of associated voxels. }
	\label{fig:3_simplification}
\end{figure*}

\begin{figure*}
	\centering
    \includegraphics[width=0.8\linewidth]{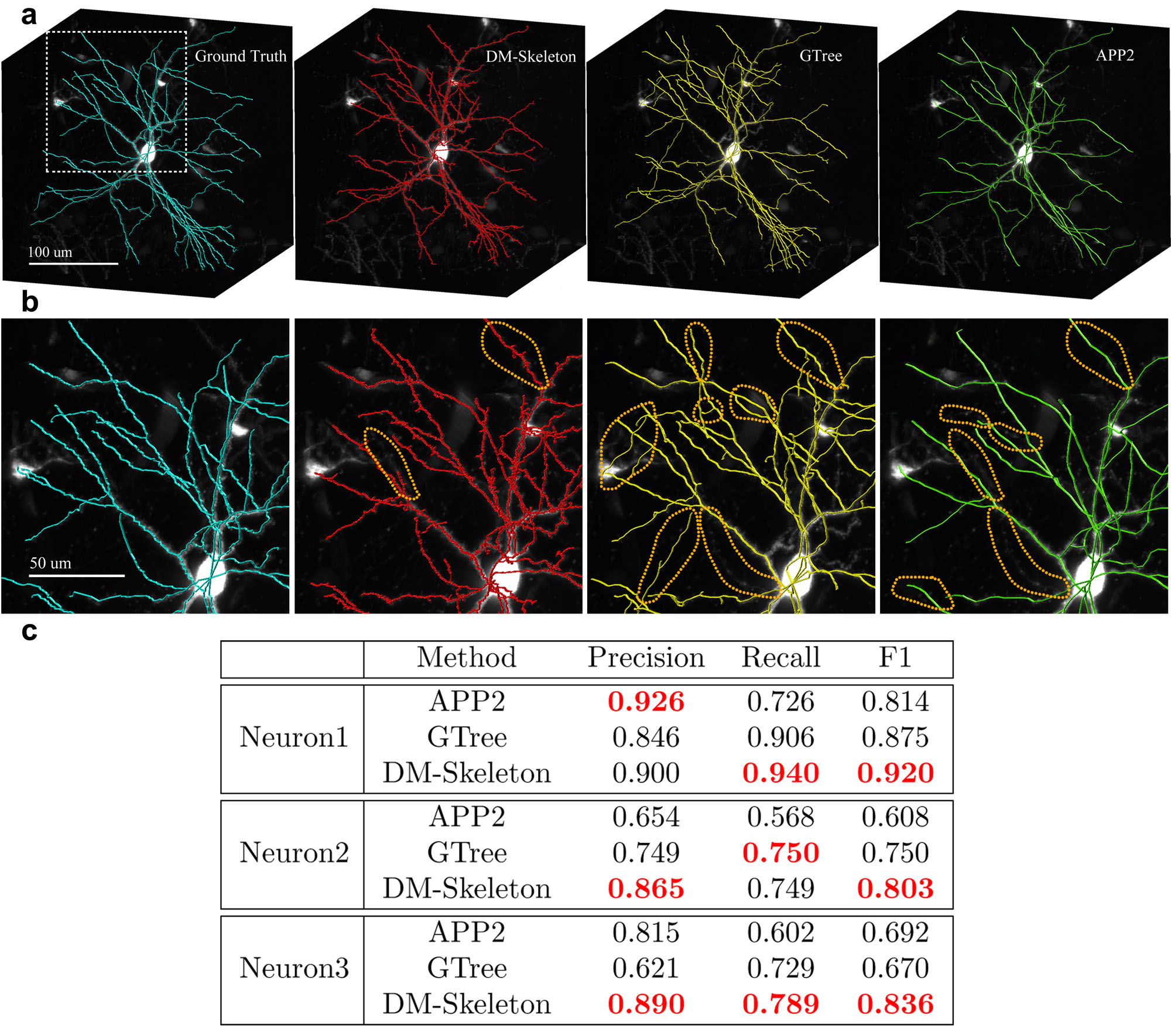}
	\caption{Results of fMOST single-neuron skeletonization. For the purpose of visualization, all thickness values were suppressed. \textbf{a,} 3D volumes of ground-truth (cyan), \myDM{} method (red), GTree (yellow) and APP2 (green) outputs. All the methods are further compared in the zoomed-in region (the region in the white square shown on the ground truth) in ({\bf b}). \textbf{b,} Ground-truth (cyan on the left), and comparisons of zoomed-in regions of \myDM{} method and GTree method. 
	APP2 output (green) clearly misses many branches and is considered having the worst performance. 
	False positives and false negatives are highlighted in orange circles. GTree method has many false positives and APP method misses many branches. Our \myDM{} method has a visible advantage over GTree and APP2 in these examples.
	\textbf{c,} Evaluation on fMOST dataset, best scores achieved are highlighted in red. \myDM{} used persistence threshold = 256 and simplification threshold = 0.2 for all the cases. The APP2 method does not have a good performance in terms of F1-score. On average, when compared with the GTree method, \myDM{} has around 14\% advantage on precision and slightly better recall; it achieves the best F1-scores ($\sim 5\%$ to $\sim 16\%$ better) in all the regions.}
	\label{fig:4_fMOST}
\end{figure*}

\begin{figure*}
	\centering
    \includegraphics[width=0.8\linewidth]{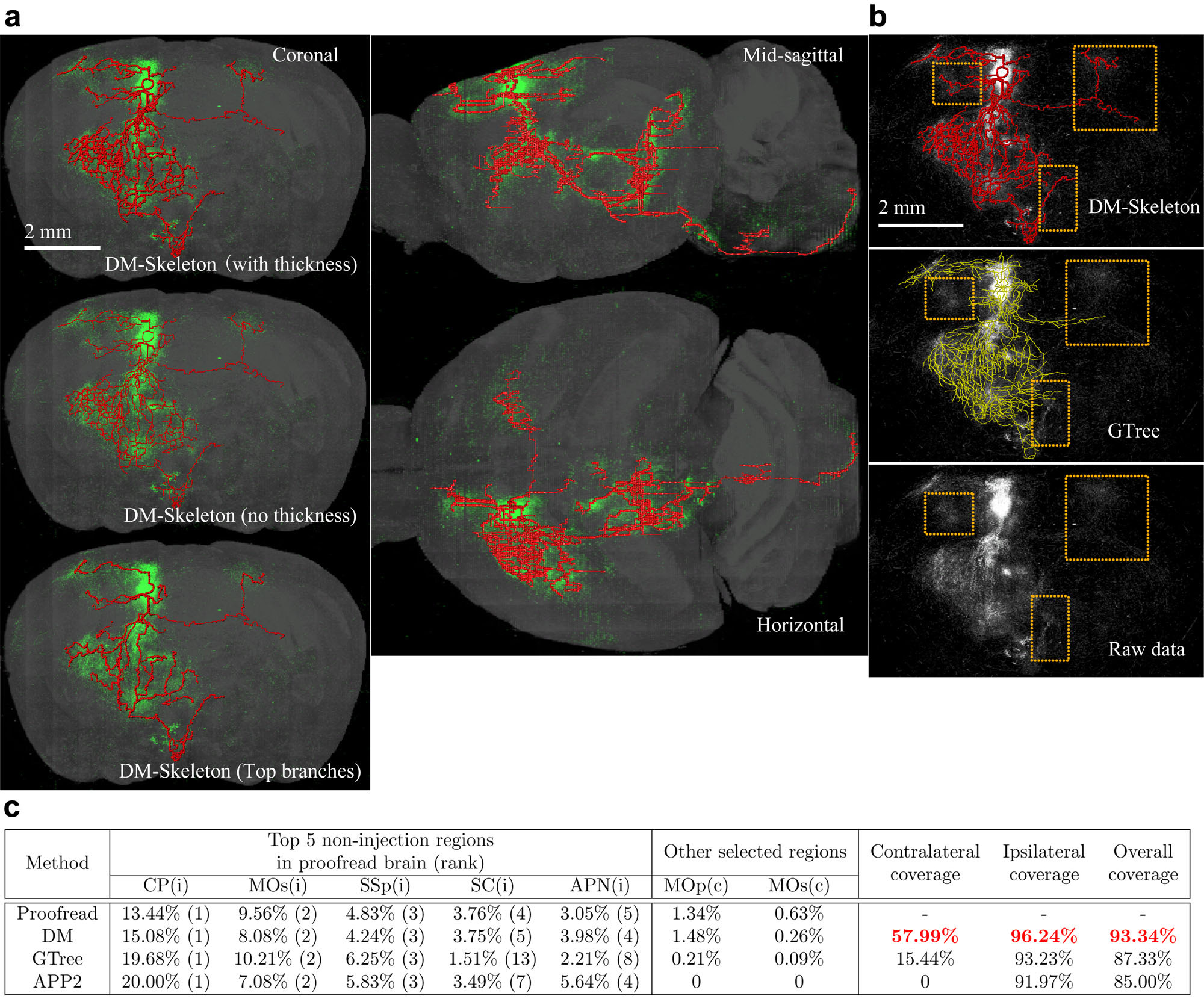}
	\caption{ Skeletonization result on STP data with tracer injection. \textbf{a,} On the left, we show the results in a coronal plane. The result with thickness assigned (top) have tree nodes with greater thickness in regions with higher process density. In the output before thickness assignment (middle), the thickness of every tree node is set to a constant. Top 20 branches are also shown (bottom). Selecting top branches based on total length can provide a concise summarization. The \myDM{} result after thickness assignment are presented from other angles on the right. The images and summaries were rescaled along the rostral-caudal direction for isotropic visualization. \textbf{b,} The APP2 outputs (shown in Supplementary Fig. \hyperref[supp_fig:stp_results_all_methods]{3}) do not have comparable quality to the other two, so the visual comparison here is between GTree and our method. The raw data (bottom) has several challenging regions marked by orange rectangles. \myDM{} method successfully captures these regions, while the GTree method misses these regions. In addition, the GTree method has many redundant edges in the high density region, which causes difficulties in observing the main skeleton. \textbf{c,} We computed the weight distribution over regions for all summarization methods and the proofread brain with an injection in the primary motor cortex (MOp). The same brain regions in either hemisphere were considered separately. The 5 projection regions with top weights are analyzed for the summarization methods. The injection site is on the ipsilateral hemisphere. The \myDM{} method recovers the correct top 5 regions, while the GTree method has relatively low ranks for SC and APN regions. In other regions, e.g., the MOp and MOs in the contralateral hemisphere, the GTree method does not capture the signals very well and the APP2 method completely misses those regions. Our \myDM{} has the most accurate summary for all cases. The abbreviated region names are Caudoputamen (CP), Secondary motor area (MOs), Primary somatosensory area (SSp), Superior colliculus (SC) and Anterior pretectal nucleus (APN). The suffixes (i) or (c) correspond to the ipsilateral or contralateral hemisphere. \myDM{} is abbreviated as DM in the table for better presentation.}
	\label{fig:5_tracer_injection}
\end{figure*}

\begin{figure*}[ht!]
    \centering
    \includegraphics[width=0.8\linewidth]{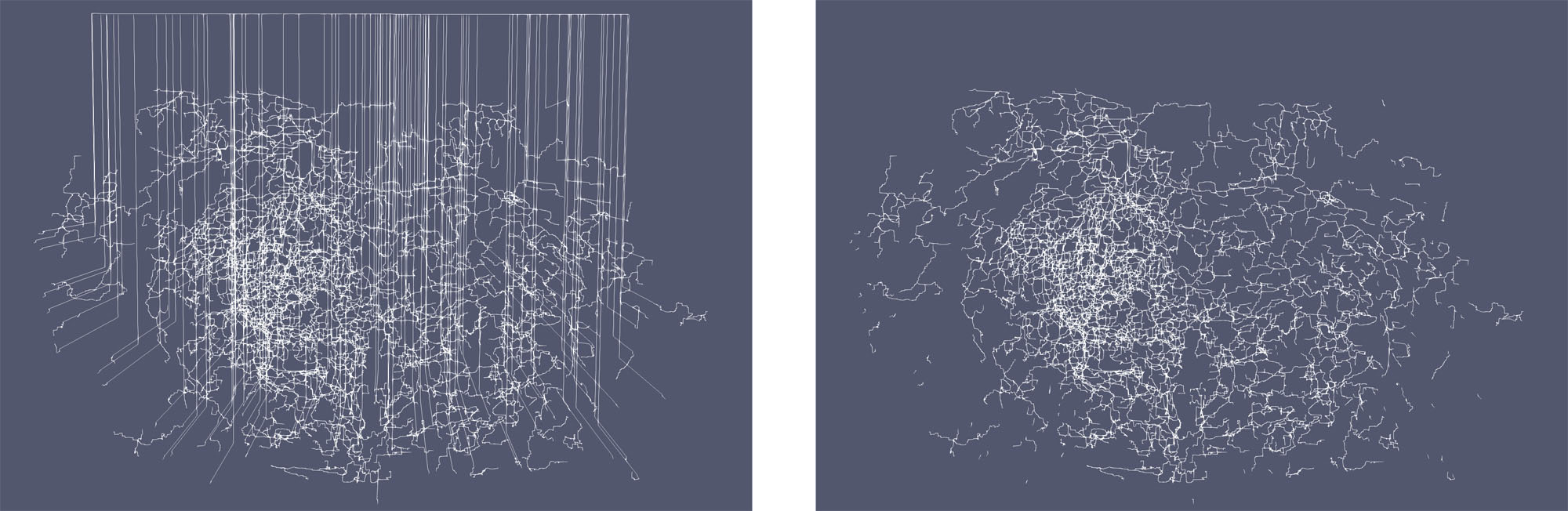}
    \caption{On the STP dataset, due to the degenerated gradient on the background pixels, there are redundant edges going to the boundary (on the left). Those edges can be filtered out by a small density threshold (result on the right) since the pixel value of the background is zero after pre-processing.}
    \label{supp_fig:stp_boundary_edges}
\end{figure*}

\begin{figure*}[ht!]
    \centering
    \includegraphics[width=0.8\linewidth]{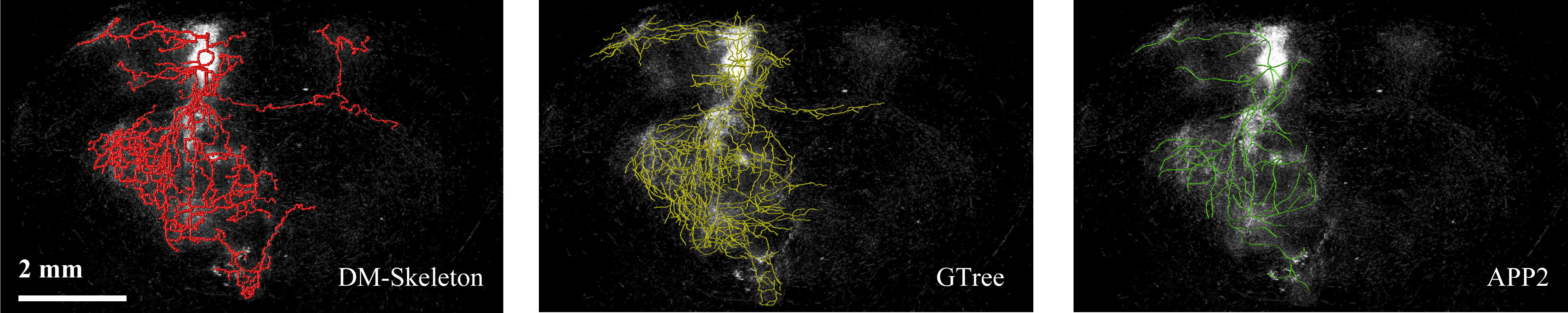}
    \caption{Summarization results of \myDM{} method (red), GTree (yellow) and APP2 (green).}
    \label{supp_fig:stp_results_all_methods}
\end{figure*}

\begin{figure*}[ht!]
    \centering
    \includegraphics[width=0.8\linewidth]{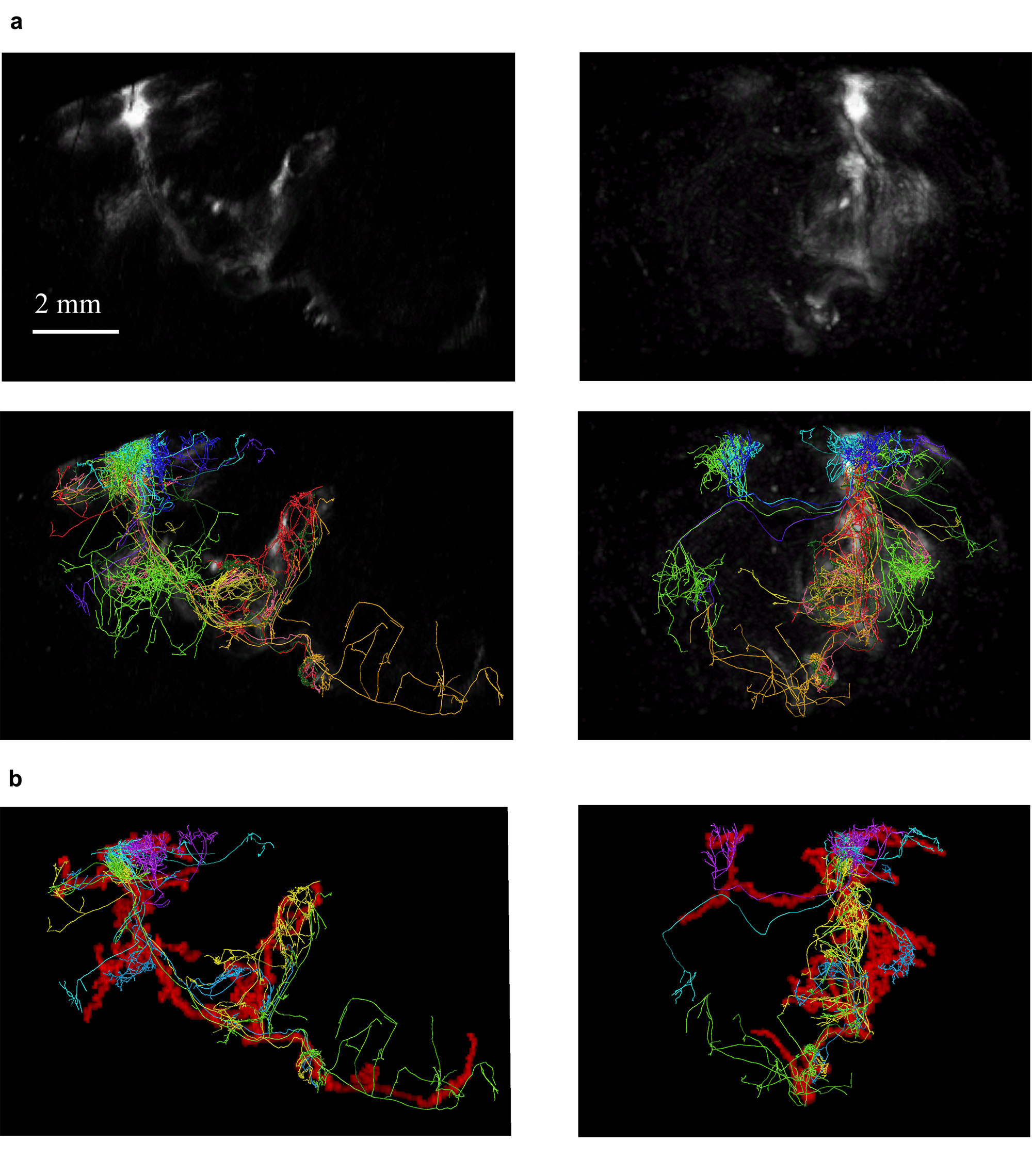}
    \caption{Mouselight superposition. \textbf{a,} The selected 11 mouselight neurons are superposed with the proofread volume and compared in 2 angles. We mapped the volume to atlas space and slightly raised the brightness and contrast for easier comparison. \textbf{b,} The neurons are also superposed with the \myDM{} output. To see the clear contour of \myDM{} summarization, we only selected a representative subset of those 11 neurons which can show the shape with lower complexity.}
    \label{supp_fig:mouselight_vs_volume}
\end{figure*}

\indent {\bf Summary method overview}
The high-level workflow of our new Discrete Morse-based (\myDM{}) pipeline is shown in Fig. \ref{fig:1_workflows}. The detailed flows for high-resolution single neuron data (e.g,  fluorescent micro-optical sectioning tomograph (fMOST)) and for whole-brain 3D densely labelled tracer injection data (e.g, serial two-photon tomography (STP)) are slightly different, as the types of input and goals are different. 
Nevertheless, both workflows have three main components, namely preprocessing, skeletonization, and simplification. (For details, see \hyperref[subsec:method_DM_pipeline]{Method} section.)\\
\begin{figure*}
	\centering
    \includegraphics[width = 0.8\linewidth]{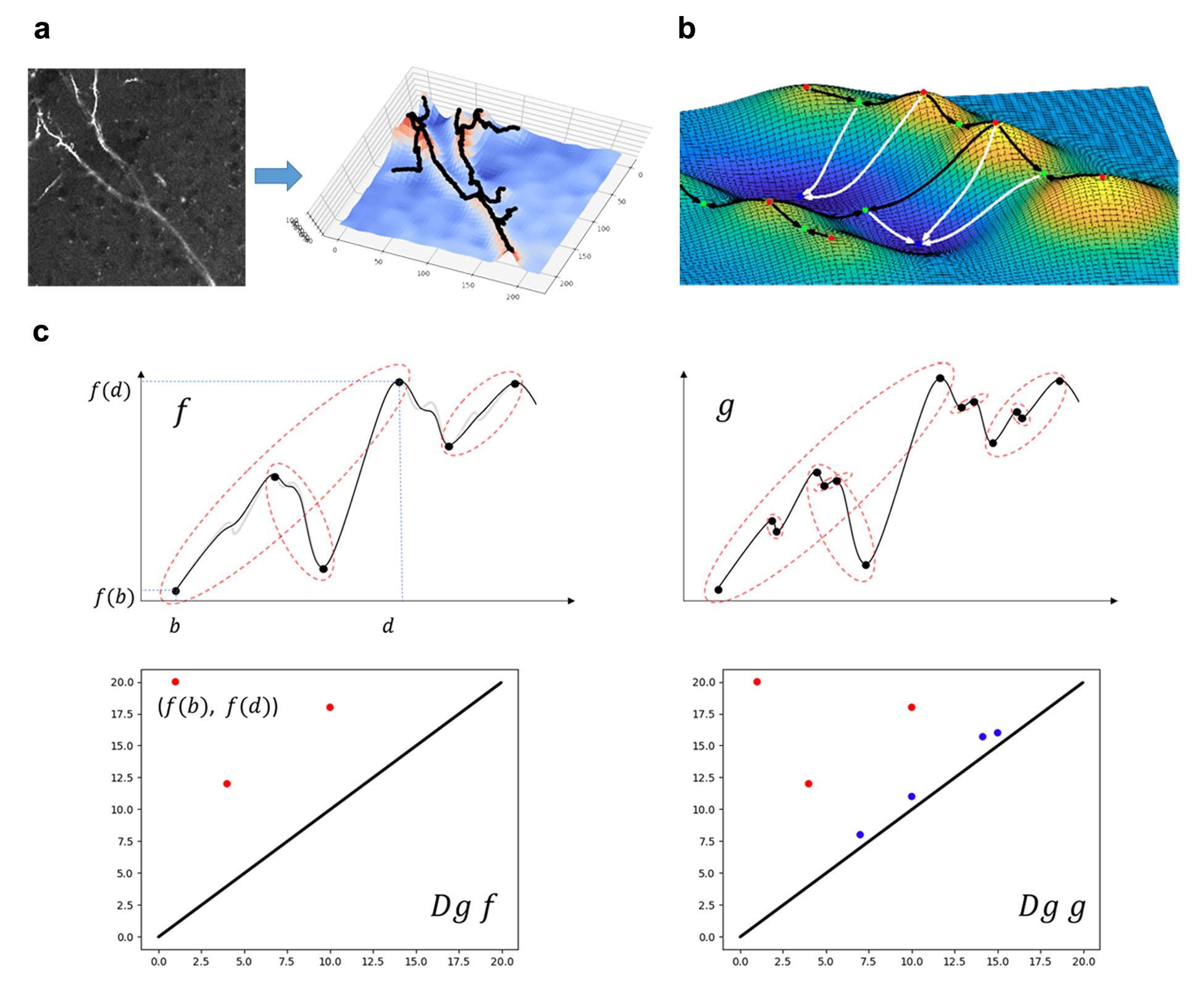}
	\caption{Illustration and basic concepts for Discrete Morse theory. \textbf{a,} An example of a 2D image (left) converted to a density function with the graph (terrain) of this function (right). The weak y-shape connection is captured by the mountain ridges (shown in black) on this terrain. \textbf{b,} An example of 2D terrain (graph of a 2D function): red points are local maxima, green points are saddles, while blue points are local minima. The white paths are some examples of the integral paths following the local gradients, ending in minima. The black curves are a collection of 1-stable manifolds (integral paths between maxima and saddles). \textbf{c,} An example of persistence pairs on 1D functions. For a function $f: \R \to \R$, topological features (connected components in this simple case) are 'born' at local minima, and 'die' at local maxima. Each persistence pair $(b, d)$ indicates the birth and death of some feature, born at $f(b)$ and killed at $f(d)$. This gives rise to a point $\left( f(b), f(d) \right)$ in the so-called persistence diagram $Dg f$ w.r.t.$f$ (lower pictures). The persistence of this feature is the difference in function values $|f(d) - f(b)|$ which can be considered as a measure of importance as it gives how ``long'' the feature persists. In the figures, persistence pairings are marked by red dotted curves. The function $g$ (shown on the right) can be viewed as a noisy perturbation of function $f$ (shown on the left). The function $f$ has 3 prominent features (persistence pairs), while the perturbed version also has additional ``smaller’’ features with lower persistence (importance). As shown in the persistence diagrams (lower row): In addition to the three red persistence points, diagram $Dg ~g$ also has blue points close to the diagonal, indicating features whose persistence (i.e., deathtime$-$birthtime) is small. These small features ("noise") can be detected using their persistence values and removed in our algorithm.  }
	\label{fig:2_DM_tutorial}
\end{figure*} 
\noindent \myDM{} takes 3D image stacks as input. In Step 1, the input image is converted to a density field defined on a 3D grid $\DensityMap : \SimpComplex \rightarrow \R$. 
For fMOST data, raw image intensity was treated as the density value subjected to Morse skeletonization. For the STP data, a process-detection step \cite{Samik_DM++} was first applied to segment labelled axon fragments in the high-resolution 2D images. A 3D volume was created to summarize the density of axon fragments within each voxel at lower resolution. This summary volume was the density field subjected to Morse analysis. Given the starting density field $\DensityMap$, the goal of the DM-skeleton algorithm is to capture center-lines passing through (relatively) high density regions. By viewing this density map as a terrain (see Fig. \ref{fig:2_DM_tutorial} for a 2D example), this step corresponds to extracting the 'mountain ridges' of this terrain. This extraction was achieved using 1-stable manifolds from discrete Morse theory. \\
\indent In Step 2, a variant of the persistence-guided discrete Morse-based framework of \cite{WWL15,DWW18} was applied to the 3D density field $\DensityMap$. We call the output the Morse 'graph skeleton' $G$. Note that this approach takes into account the global topology of the density field: the 1-stable manifold connects through low-density regions (e.g., gaps and weak signals around the Y-junction in Fig. \hyperref[fig:2_DM_tutorial]{2a}) reliably. 
Finally, in Step 3, we extracted either a forest (e.g, each tree representing a single neuron in fMOST data) or a tree summary from the Morse graph skeleton $G$. The forest output may have false positives, thus we developed a simplification module to further remove such branches (see Fig. \ref{fig:3_simplification}; more details in Method). \\
\indent {\bf fMOST sparse labeling dataset.} 
Three semi-manually reconstructed neurons from fMOST dataset were taken as ground truth.  A $224 \times 224 \times 251 \upmu $m 
neighborhood around the soma was taken as the test dataset. More details about the test dataset and the ground truth are provided in \hyperref[subsec:method_data_collection]{Methods}.
\label{res:fMOST}
The Morse graph skeleton obtained from the persistence-guided discrete Morse-based framework (Step 2) was fed to the simplification module (Step 3). 
In Step 3, the positions of somata were taken as given and used as roots to extract a forest such that one tree is rooted at each soma.
Note that there exist many automatic methods for reliably detecting soma positions \cite{SomaDetection_Yan_2013, SomaDetection_Cheng_2016}. Automatic methods can be applied to reduce human labor when a region has many somata. 
The tree rooted at the soma of the target neuron was then simplified. In this step, each tree node was assigned a score based on density (see \hyperref[supp:cal_scores_smoothing]{Supplementary Methods}). Next, the \rootgrower strategy was applied (see \hyperref[subsec:method_DM_pipeline]{Method} section) to grow a simplified neuron tree. 
This strategy (instead of \leafburner) is more suitable for the fMOST dataset because there is observable autofluorescence scattered in the extracted regions. A bottom-up process (\leafburner) would stall in such regions when the autofluorescence has high-intensity values similar to the somata. \\
\indent We evaluated the performance of the \myDM{} pipeline on three fMOST single-neuron regions. We compared our output with the state-of-the-art approaches GTree \cite{Hang_GTree2018} and APP2 \cite{APP2_Hang2013} methods. For GTree and APP2, we tested different parameters and used the outputs having the fewest gross errors for comparison. We compared the precision, recall and F1-score 
of the outputs of these three methods over ground-truth reconstructions (Fig. \hyperref[fig:4_fMOST]{4c}). 
In all three regions, we note that both our \myDM{} and the GTree method performed better than the APP2 method. Furthermore, on average, our \myDM{} method outperformed GTree by around 14\% in precision and 3\% in recall. This suggests that our \myDM{} has significantly fewer false positives than the output of GTree. \\
\indent To further illustrate the potential use of our output, we proofread the outputs from GTree and \myDM{}  using 3D virtual finger tool \cite{Peng_virtual_finger_2014} on fMOST regions. 
Proofreading was performed on both the GTree outputs and the \myDM{} outputs by two independent annotators without domain-specific knowledge and with no prior knowledge of the ground truth. 
The annotators also constructed the skeleton from scratch on the region of neuron1 using the same proofreading tool with an average time spent of roughly one hour. Supplementary Table \ref{table:proofreading_fMOST} shows that the automatic methods reduce the human time of annotation by at least two thirds. 
Precision, recall, and F1-score were calculated to compare proofread results with the ground truth (Supplementary Table \ref{table:proofreading_fMOST}).  All of these metrics demonstrate that proofread results even by naive annotators were consistent and of high-quality. Importantly, the time spent proofreading the \myDM{} outputs was much shorter than the time spent proofreading GTree outputs. The reason is that \myDM{} outputs only have minor early-termination issues, which can be corrected by extending corresponding traces. By contrast, GTree outputs usually have a considerable amount of false positives. They also occasionally miss important mainstream branches, which causes difficulties and prolongs the proofreading process. \\
\indent To utilize our framework, one labeled dataset was used to tune the parameters (for persistence simplification and tree simplification). These parameters were then applied to other datasets with similar contrast and SNR. In the results reported above, Neuron-1 from the fMOST dataset was used to tune the parameters. We chose multiple persistence and tree simplification thresholds. For each such pair of choices, we generated the final tree and computed its F1-score compared with ground-truth tree (Supplementary Table \ref{table:choose_persistence_simp_threshold}). We then selected the thresholds that give the best F1-score and apply the same thresholds to all other fMOST neurons. From Supplementary Table \ref{table:choose_persistence_simp_threshold}, we note that results were reasonably stable to perturbations of thresholds.\\
\indent {\bf STP tracer injection dataset.}
\label{subsec:STPpreprocess}
When applying the \myDM{} pipeline on the tracer injection dataset, 
in Step 1, the original image data 
(see Section \hyperref[subsec:method_data_collection]{Data Collection}) 
was passed through a hybrid deep CNN with topological priors\cite{Samik_DM++} to detect axon fragments. The output of this automated detection process was then manually proofread and corrected. The proofread images were summarized into lower resolution to construct a manageable 3D volume of the entire brain. 
The remaining process is similar to the single neuron skeletonization algorithm above. However, there are some notable differences. On one hand, the pre-processed STP data was clean and did not have as much noise as the autofluorescence and bright background in fMOST data, making the \leafburner simplification strategy preferred over the strategy \rootgrower for STP dataset (see Fig. \ref{fig:3_simplification}, and also \hyperref[subsec:method_simplification]{Method} section). On the other hand, the tracer injection data has more discontinuities and gaps. Therefore, a Gaussian filter (kernel radius 2) was applied to the image stack (after pre-processing) before further processing. 
The persistence threshold was adjusted to account for bit-depth differences. The simplification threshold for the tracer data was taken to be the same as for the single neuron data set even though we changed the strategy for simplification. This parameter was robust to input image types because an average score of all tree nodes was used as the reference (\hyperref[supp:cal_scores_smoothing]{Supplementary Methods}).\\
\indent The Morse graph skeleton obtained in Step 2 may have some redundant straight segments connecting signals to the boundary of the domain (Supplementary Fig. \hyperref[supp_fig:stp_boundary_edges]{2}). This boundary effect is caused by the zero-value background and degenerated gradient on those pixels in the cleaned STP data. Such redundant segments can be easily removed based on distance to the nearest non-zero voxel. \\
\indent We can in principle further streamline the output using flow vectors estimated with weighted principal component analysis \cite{Vector_simp_Delchambre_2014}. Paths not well aligned with estimated flow vectors can be removed as long as doing so does not increase the number of connected components that make up the output. In the results shown below, we did not apply this flow vector simplification, but results with or without this flow vector simplification step are compared in Supplementary Fig. \hyperref[supp_fig:stp_vec_simplification]{4}. The output with flow vector simplification is visually cleaner and still has good coverage rates. \\
\indent In Step 3, the root position was manually selected within the projection site. Then, a shortest path spanning tree was extracted and simplified. In contrast with the single-neuron skeletonization, the tracer injection skeleton has a {\it weight} assigned to each tree node after simplification. The weight of a given tree node $v$ represents the total weight of voxels whose nearest neighbor in the tree node set is $v$ (see Fig. \ref{fig:3_simplification}, and also \hyperref[subsec:method_simplification]{Methods} section). This is a weight-preserving process where the weights of the tree nodes of the output preserve the total tracer density of points assigned to that node.\\
\indent In addition, for visualization purposes we assigned a thickness value for each tree node. The thickness value uses the number of voxels instead of their weights to avoid excessively large tree nodes around the injection site (see Fig. \hyperref[fig:5_tracer_injection]{5a} and \hyperref[subsec:method_simplification]{Methods} section).
Keeping only top branches (based on length) can provide a more concise summary or skeleton. See Fig. \hyperref[fig:5_tracer_injection]{5a}. \\
\indent Visually, the output by APP2 method is over-simplified and misses many obvious signals (Supplementary Fig. \hyperref[supp_fig:stp_results_all_methods]{3}). 
GTree and \myDM{} method have more plausible results (Fig. \hyperref[fig:5_tracer_injection]{5b}). 
In Fig. \hyperref[fig:5_tracer_injection]{5b}, we further compared GTree and \myDM{} outputs in several regions. From the raw data, we can see that these regions, corresponding to motor cortex and striatum in the contralateral hemisphere, contain visible signals. GTree, however, does not have enough branches covering these regions. This may be because GTree is strongly dependent on brightness and contrast, and is not robust enough to handle regions of images with relatively low brightness. This necessitates fine-tuning of the pixel-value histogram prior to running GTree. In contrast, the Discrete Morse-based method is able to overcome these issues because only the relative order of pixel-values matters and effectively there is a built-in histogram localization. Thus the method is automatically adaptive to different types of input images, as well as to different local pixel intensity distributions, with minimal image pre-processing requirements. 
We also note that even the top few branches from our \myDM{} output (the image at the left bottom corner in Fig. \hyperref[fig:5_tracer_injection]{5a}) still capture key branches and present a good coverage of different brain regions (including those within orange boxes shown in Fig. \hyperref[fig:5_tracer_injection]{5b}, which are missed in the outputs by GTree and APP2 methods). \\
\indent To verify the biological interpretability of the skeletonization results, we first mapped the 3D volume of the simplified tree to the Allen mouse brain atlas \cite{AllenMouseBrainAtlas} for regional projection strength analysis (\hyperref[supp:regional_analysis]{Supplementary Methods}). The intensity of projected pixels reflected the weights preserved by tree nodes. The projected segments were categorized into the corresponding brain regions based on the atlas map. We applied the same procedure to the proofread volume. A good summarization should cover most of the signals in the input data. Thus, the regions covered by the skeleton-based summary and the conventional connectivity matrix based summary should coincide. Moreover, the distributions of weights over different regions should be similar.
In Fig. \hyperref[fig:5_tracer_injection]{5c}, we computed the percentages of (projected) weights within certain regions over the total weights for the proofread data and all three summarizations. Note that the output from GTree is merely a skeletonization without thickness. \\
\indent We selected the top 5 regions with the highest weights based on the conventional connectivity matrix summary and found the ranks of these regions in the summarization outputs.  \myDM{} output has the identical top 5 regions as those calculated from the proofread data, while GTree or APP2 have difficulties in capturing the signals in SC and APN. The \myDM{} output captured the projection into the contralateral areas, while 
the GTree output did not cover many of these regions. The APP2 output completely missed the contralateral hemisphere. There are other regions where our weighted tree qualitatively matches the conventional summary, while GTree and APP2 have weight percentages close to zero.
The coverages for each hemisphere as well as the entire brain were calculated individually. Our \myDM{} had the best coverage rates in all cases. In the contralateral hemisphere, it outperformed the GTree method by more than 40\%. There is a small discrepancy in the weight distribution over regions between the conventional process density summary and the \myDM{} based summary. The reason is that \myDM{} did not capture some poorly labelled regions, and the voxels may be assigned to a tree node in a neighboring region.\\
\indent In addition, we compared our \myDM{} results with single neuron data from the Mouselight project \cite{Mouselight}. Eleven neurons which have soma positions in the injection site of the STP dataset were selected and quantitatively summarized. Visually, the \myDM{} output showed good correspondence with this collection of neurons, which were expected to cover similar regions as the STP volume (Supplementary Fig. \hyperref[supp_fig:mouselight_vs_volume]{5}). 

\section*{Discussion} \label{sec:Discussion}
\indent In this paper we introduced a new conceptual framework of data analysis utilizing Discrete Morse Theory for extracting underlying tree structures and applied it to two neuroscience problems of current importance: the automatic extraction of single neuron skeletons, and biologically meaningful analysis of mesoscale connectivity mapping data. Apart from being conceptually and mathematically well grounded, this approach significantly outperforms state-of-the-art approaches relevant for these problems. \\  
\indent Current approaches to analyzing brain-wide tracer-injection data all start from regional connectivity matrices, which lose the biologically important tree-like structure of neurons composing those projections. Apart from respecting the collateral branching patterns of the underlying set of neurons and providing a natural bridge to single neuron data, the \myDM{} method utilizes the global structure of the signal and is robust to noise and missing data, and naturally adaptive to signal intensity variations. We demonstrated the approach using a tracer-injected whole-brain data set. The resulting tree-skeleton faithfully captured the regional "connectivity matrix" information, and significantly outperformed baseline methods. In addition it captured the collateral branching patterns, consistent with corresponding single neuron skeletons with somata in the injection region. We expect that this approach will be useful in the future for mesoscale brain connectivity mapping using tracer injections and form a bridge to the single neuron data.\\
\indent We also demonstrated significance performance improvements for single neuron reconstruction over existing best in class methods. Automated single neuron skeletonization is a well-studied problem with a large associated literature and hundreds of algorithms. Our methodology breaks new ground by taking a different conceptual approach that leads to a simple and theoretically transparent algorithmic framework simultaneously with performance improvements. As in the case of tracer-injection skeletonization, the method takes into account the global structure of the data, is robust to noise and missing data, and is data-adaptive as construction of the Morse skeleton depends on the order relations between neighboring pixel intensities rather than their absolute values.\\
\indent While our method achieves strong performance, it is computationally expensive compared with other methods. The computational bottleneck comes from the persistence-guided discrete Morse-based framework. Computing persistence pairings takes $O(n^3)$ running time in the worst case (although usually significantly faster in practice), where $n$ is the number of cells that make up the input cell complex. To address this bottleneck, we utilized DIPHA \cite{DIPHA_Bauer_2014} to compute persistence pairings. This algorithm is distributed and allows for a significant speedup compared to non-distributed persistence algorithms. Further code optimizations over our current implementation are possible and run-time can be further reduced in future work. 
Despite the higher computational complexity, the conceptual elegance and theoretical transparency, performance improvements including significant reduction in human proof-reading times, and incorporation of prior biological structure are strong arguments in favor of the approaches proposed here. 

\section*{References}

\bibliographystyle{naturemag}
{\footnotesize
\bibliography{references} 
}


\section*{Data and code availability}
The fMOST and STP data were collected as a part of Brain Initiative Cell Census Network and shared online. 
The raw images of the STP dataset are available from: \url{ftp://download.brainimagelibrary.org:8811/biccn/huang/connectivity/anterograde/180830_JH_WG_Fezf2LSLflp_CFA_female_processed/}.
The raw images of the fMOST dataset are available from: \url{ftp://download.brainimagelibrary.org:8811/biccn/zeng/luo/fMOST/732664811/}. Single neuron reconstruction ground truth are available from \url{ftp://download.brainimagelibrary.org:8811/biccn/zeng/luo/fMOST/cells/732664811/}. 
Processed projection summary of STP data, subregions of fMOST data, the code and documentation are available on \href{https://github.com/wangdingkang/DiscreteMorse}{Github} at \url{https://github.com/wangdingkang/DiscreteMorse}.

\section*{Acknowledgements} \label{sec:Acknowledgements}
The authors thank Katherine Matho and Kathleen Kelly for helpful discussions on the test datasets. 

\noindent This work is in part supported by National Science Foundation under grants CCF-1740761, RI-1815697, and DMS-1547357, National Institute of Health under grant R01-EB022899, MH114821 and MH114824. We would also like to thank Crick-Clay Professorship, Mathers Charitable Foundation and H N Mahabala Chair, IIT Madras for their support. 

\section*{Author contributions} \label{sec:Author_contributions}
D. W., L. M., S. W., Y. W. and P. M designed the pipeline. D.W. and L.M. implemented the pipeline. B. H., S. B., J. J., K. R., X. L. and M. L. provided the test datasets. D. W. and L. M. collected results on all test datasets, and together with Y. W. and P. M. analyzed the results on fMOST dataset. D. W., B. H., X. L., Y. W. and P. M. analyzed the results on STP dataset. D. W., L. M., B. H., S. B., J. J., X. L., Y. W. and P. M. wrote the manuscript. 

\section*{Competing interests} \label{sec:Competing_interests}
The authors declare no competing interests.

\clearpage
\pagebreak
\newpage
\pagestyle{plain}
\section*{\Large Methods} \label{sec:Methods}

%

\subsection{Data Collection.}
\label{subsec:method_data_collection}
The Serial Two-Photon (STP) dataset presented in this paper was collected as a part of Brain Initiative Cell Census Network \cite{Matho_BICCN}. Specific Cre-dependent transgenic mouse lines were crossed with IslFlp reporter lines. Flp-dependent AAV tracers were utilized to reveal cell type-specific axon connection \cite{Neural_circuits_JoshHuang2013}. 
Each brain was prepared and imaged using STP tomography \cite{Serial_two_photon_Ragan2012} with $1 \upmu \text{m} \times 1 \upmu \text{m}$ in-plane resolution, and sectioned coronally every 50 $\upmu \text{m}$. Two channels of 16-bit data were collected, where Channel 1 collected the autofluorescence and Channel 2 collected the fluorescent tracer information.  Only Channel 2 data were used in the subsequent analysis. One STP dataset was involved in the development and demonstration of methods in this paper (available from: \url{ftp://download.brainimagelibrary.org:8811/biccn/huang/connectivity/anterograde/180830_JH_WG_Fezf2LSLflp_CFA_female_processed/}). 

\noindent The dual-color fluorescent micro-optical sectioning tomography (fMOST) data presented in this paper , both raw images and single neuron reconstruction data (“ground truth”), were collected as a part of Brain Initiative Cell Census Network and downloaded from Brain Image Library (available from \url{http://www.brainimagelibrary.org/}). 
\noindent Specific Cre-dependent transgenic mouse lines with reporter transgenic mouse lines. Sparse labeling of the entire neuron was achieved by injecting small amount of viral tracer. The brain was fixed and embedded in resin before mounted for fMOST imaging at $0.32\upmu \text{m} \times 0.32\upmu \text{m}$ in-plane resolution and sectioned serially at $1\upmu \text{m}$ . Two channels of data were collected, with 16-bit data in each channel \cite{Dual_color_imaging_Gong2016}. Only green channel contained the neuron tracing information and therefore used in the current study. One fMOST dataset was involved in the development and demonstration of methods in this paper (available from: \url{ftp://download.brainimagelibrary.org:8811/biccn/zeng/luo/fMOST/732664811/}). Single neuron reconstruction was performed using a combination of automatic tracing and manual annotation \cite{TeraVR_Wang2019}. Information of nodes and edges of individual neuron was stored in separate files in swc format (available from: \url{ftp://download.brainimagelibrary.org:8811/biccn/zeng/luo/fMOST/cells/732664811/}). 

\subsection{Data pre-processing.}
\label{subsec:method_data_preprocessing}
For fMOST data, neighborhoods of 3 reconstructed neurons were taken as the test datasets. Each neighborhood volume was saved as one VTK file (simple legacy format). The associated single neuron reconstruction data, saved in SWC format, were taken as ground truth for the subsequent analysis. The VTK and SWC files are available on \url{https://github.com/wangdingkang/DiscreteMorse}.

The STP dataset was first processed with a topologically motivated convolutional neural network \cite{Samik_DM++} for the detection of the tracers. The network, termed as DM++, takes in whole STP sections and divides them into $512 \times 512$ pixel tiles. These tiles are passed through a topological algorithm based on Discrete Morse \cite{DWW17} and a CNN counterpart for determining the topological and neuronal priors, respectively. The topological priors capture the faint connectivity which is used to boost the performance of the CNN in a supervised Siamese setting for the dual priors incorporated into the DM++ framework. The final output likelihood map is converted into a binary mask for the neuronal processes using an optimal empirically determined threshold. This captures most of the processes in the tiles, which are then stitched back together to form a mask for an entire reference section of the brain. 

\noindent The preliminary outputs of process detection were manually verified for the entire brain by an experienced neuroanatomist using Fiji \cite{FIJI_Schneider2012}.  Briefly, the preliminary outputs consisting of detected signal were masked with the original brain section image and error corrected using the pixel painting tool. The filled processes were identified as those having a brighter intensity compared to the background.

\noindent The proofread brain from the previous step was annotated in the format of binary images. The images were further downsampled to the desired resolution using the sum method, which preserved all the signal information. Volumes of neuronal axons were summarized by counting the number of pixels ($1\upmu \text{m} \times 1\upmu \text{m}$ resolution) containing fluorescent signal into $10\upmu \text{m} \times 10\upmu \text{m}$ in coronal sections while keeping the inter-section spacing the same ($50\upmu \text{m}$). The resulting images are 8-bit, and the value for each pixel indicates the signal strength for its corresponding voxel. The image stack was saved in VTK simple legacy format and is available on \url{https://github.com/wangdingkang/DiscreteMorse}.

\subsection{DM-Skeleton}
\label{subsec:method_DM_pipeline}
This section provides intuition and descriptions for the \myDM{} pipeline; recall the workflow in Figure \ref{fig:1_workflows}. 
\subsection{Step 1: Pre-processing}
Each image volume in VTK format was loaded as an image stack, and converted into a density field  $\DensityMap : \SimpComplex \rightarrow \R$ defined on the cubical complex (3D grid) $\SimpComplex$, where each vertex corresponds to a voxel in the input image and has a density value. 
If the input is fMOST data, then the density value at each voxel is simply the pixel value in the input raw image stack. 
For whole-brain tracer injection STP data, a significant portion of the raw images is background (see the example in Figure \ref{fig:1_workflows}). 
Hence we first apply the learning-based process-detection module \cite{Samik_DM++} to remove the background and segment the foreground. 
We further apply a Gaussian filter to smooth the values across the domain. 

\subsection{Step 2: Skeletonization.}
\label{subsubsec:step2} 
The Discrete Morse graph reconstruction algorithm \cite{WWL15,DWW18} takes a density field of any dimension as input, and outputs a graph skeleton capturing center-lines passing through relatively high density regions. 
In our case, the input density field $\DensityMap : \SimpComplex \rightarrow \R$ is defined at vertices of a cubical complex $\SimpComplex$ of the domain. In all subsequent operations, only the 2-skeleton of this cubical complex $\SimpComplex$ is needed, that is, we assume $\SimpComplex$ consists of vertices, edges, and squares. Cubic cells are not needed in the algorithm.  

\noindent To explain the main idea, consider first the smooth case where we have a smooth function $\DensityMap: \Omega \to \R$ over the domain $\Omega$.  
Consider the terrain of the density function values plotted over the domain (Fig. \ref{fig:2_DM_tutorial}) where the terrain of a function defined on $\R^2$ is given. 
The underlying graph skeleton of $\DensityMap$ can be captured by the mountain ridges of this terrain (Fig. \hyperref[fig:2_DM_tutorial]{2a}). Locally, the density along these mountain ridges is higher than the density off of them. 
These ridges form the so-called \emph{1-stable manifold} of the function in Morse theory, and are defined by the integral lines ``connecting'' local maxima to saddle points (Fig. \hyperref[fig:2_DM_tutorial]{2b}). 
(An integral line is a curve in the domain where at any point on it, its tangent vector coincides with the gradient of the density field. Integral lines are thus intuitively flow lines, following the steepest descending direction of the density fields.) 

\noindent Inside the algorithm, roughly speaking, ridges are associated with certain persistence values, as captured by the so-called persistent homology \cite{Edelsbrunner2002}, which can be interpreted as an importance score.  This makes it easy to filter out ridges of low importance, which are typically associated with noise, from the final output by providing the algorithm with a persistence threshold. An example of simplification for a 1D function is shown in Fig. \hyperref[fig:2_DM_tutorial]{2c}.  

\noindent For the input to our algorithm, we have a density field $\DensityMap: \SimpComplex \rightarrow \R$ defined at vertices of a cubical complex $\SimpComplex$ of the domain $\Omega$ (a 3D region). 
Following \cite{DWW18}, discrete Morse theory \cite{Forman_DM_1998} is used to capture the mountain ridges mentioned above, combined with the persistence algorithm to measure importance. See \hyperref[supp:DM_implementation]{Supplementary Methods} for a description of the algorithm. To improve the efficiency of the algorithm, we modified the algorithm of \cite{DWW18} so that it works directly with cubical complexes and also uses DIPHA \cite{DIPHA_Bauer_2014} to compute persistence pairs in a distributed manner. 

\noindent These mountain ridges cover the neural branches as locally, points along these neural branches tend to have relatively higher density (signal strength) than off the branches. The global nature of the 1-stable manifolds makes the output skeleton robust to small gaps in signal, and effective at capturing junctions; see e.g., Fig. \hyperref[fig:2_DM_tutorial]{2a}, where the global nature of 1-manifolds connects through low-density region around the Y-junction. 

\noindent In ideal circumstances, we would find a persistence threshold that would remove all of the noise and only keep the ridges that make up the true neuron tree. However, because of the noisy nature of biological data and also the Discrete Morse graph reconstruction algorithm will not necessarily output a tree, we cannot take the algorithm's output as a final output. Instead, we first run the algorithm with a low persistence threshold such that we do not remove any ridges that would be part of an ideal output. Then we simplify the Morse graph skeleton in the next step.

\subsection{Step 3: Simplification.}
\label{subsubsec:step3}
The output of the above persistence-guided Morse-based framework is a geometric graph $G$, also referred to as the Morse graph skeleton. Next, $G$ is converted into a spanning forest $F$. Prior to this, both boundary and estimated flow vector simplification can be applied to $G$ to limit the number of unnecessary branches in the spanning forest. (The estimation of flow vector at each graph node can be found \hyperref[supp:initialvecsimp]{Initial vector-based Morse graph simplification} in Supplementary Methods.) 
Further simplification strategy to remove false positives and an option to control the level of details presented in the final summarization.

\subsection{Extraction of a forest.} 
The Morse graph skeleton $G$ extracted in Step 2 already serves as a good initial skeleton. Each arc $e \in E$ in graph $G = (V, E)$ is realized by a polygonal path, consisting of edges from the input grid $\SimpComplex$. 
Then, a forest (a collection of rooted trees) $F = \{T_0, ..., T_m\}$ from $G$ (See Fig. \hyperref[fig:3_simplification]{3a}) is extracted, which is in fact a spanning forest of $G$ (that is, $F$ contains all vertices from $V$, and edges of $F$ are from $G$). Here we assume we are given the set of tree roots $R = \{v_0, ..., v_m\}$. In our experiments, the positions of roots correspond to soma locations and injection sites for fMOST and tracer injection data, respectively. The injection site and soma positions can be provided by human annotators or automatically detected \cite{SomaDetection_Yan_2013, SomaDetection_Cheng_2016}.  

\noindent With $G$ and roots $R = \{v_0, ..., v_m\} \subset V$, a weighted shortest path spanning forest algorithm is applied to produce the spanning forest $F$. Intuitively, the weight of an edge depends on the density values of its endpoints, and edges in the regions of higher density values should have larger weights (i.e., smaller distances). The details of the weighted shortest path spanning forest algorithm is introduced in  \hyperref[supp:weighted_shortest_spanning_forest]{Supplementary Methods}. 
We remark that it may also be reasonable to use the minimum spanning forest of $G$. However, the shortest path spanning forest mimics the natural process of tracers spreading from the injection site, and has a better empirical performance than minimum spanning forest. 

\subsection{Tree simplification.} 
\label{subsec:method_simplification}
Now consider any tree $T = (V_T, E_T)$ from the initial spanning forest constructed above. 

\noindent The simplification based on persistence in Step 2 can remove false-positive branches to some extent, however, false positives with small density surrounded by background can still remain in each tree $T$. We further develop the following simplification strategy to remove these false positives. 

\noindent First, scores $\SimpFinalScore(v)$ are assigned to each tree node $v$ in tree $T$; the details of calculating node scores are explained in \hyperref[supp:cal_scores_smoothing]{Supplementary Methods}. 
Next, one of the following two strategies is chosen to simplify $T$ based on the $\SimpFinalScore$ scores. The first strategy, called \leafburner{}, starts from the leaves of $T$ and iteratively removes leaves with scores at most $\SimpThreshold$ to obtain the final simplified tree $\SimplifiedTree = (V_s, E_s) $. An originally internal tree node $v \in T$ will become a leave if all of its children are already removed by \leafburner{} in previous iterations. The second strategy, called \rootgrower{}, instead expands a simplified tree $\SimplifiedTree$ from the given root $\TreeRoot$ and gradually includes tree nodes from $T$ with scores at least $\SimpThreshold$. At any moment, only children of those nodes already included in the simplified tree from earlier iterations will be considered. 
\noindent Each strategy has its own merits and disadvantages. For \rootgrower{}, gaps (weak connections) might cause breaks, while \leafburner{} can be potentially stuck at branches in remote noise with high scores and fail to remove them (See Fig. \hyperref[fig:3_simplification]{3d}). 

\subsection{Weight-preserving summarization}
In this process, our goal is to summarize voxels in the input proofread brain while conserving weight information. The weights of voxels in a certain region should be assigned to a tree node of the \myDM{} output in the same region. We use a similar idea when assigning scores in the simplification step, but now only use the voxel weights. The summarization weight of a tree node $v$ is total weights of voxels whose nearest neighbor in the tree node set is $v$. Additionally, there is an upper bound $\WeightSummaryDistCap$ for the distance allowed. More specifically, $summary(v) = \sum_{x \in I, d(x, v) = d(x, V_s),d(x, v) < \WeightSummaryDistCap} \DensityMap(x)$. The upper bound $\WeightSummaryDistCap$ is intentionally large to avoid missing weights. For the STP dataset, we used $\WeightSummaryDistCap = 300 \upmu$m.

\subsection{Additional features.}
For visualization purpose, a "thickness value" can be assigned to each tree node in accordance with the SWC format (Fig.  \hyperref[fig:3_simplification]{3e}). When calculating the density score for each tree node in the simplified tree $T_s$ (\hyperref[supp:cal_scores_smoothing]{Supplementary Methods}), voxel information is leveraged in the surrounding area of that tree node. Here a similar idea is applied, we associate each voxel in the segmented foreground (in the case that the input is not preprocessed in Step 1, such as the fMOST data, then we use a low threshold to remove background)
to its nearest neighbor in $V_s$ within distance $\ThicknessDistCap$ which is specified later. The thickness value of a node $v$ is proportional to the square root of the number of associated voxels. More specifically, $r(v) = c \cdot \sqrt{N(v)}$, where $c$ is a certain constant, $r(v)$ is the radius (thickness) assigned to node $v$, and $N(v)$ is the number of associated voxels. Any tree node with no voxel associated is set to have the minimum non-zero raidus, i.e., $c$. Empirically, the default value of $c$ is $1$. In the experiment on STP dataset, we set $\ThicknessDistCap = 20 \upmu$m for the best visualization (Fig. \hyperref[fig:5_tracer_injection]{5a}). 

\noindent An option for users to control the desired level of detail is provided. In particular, given the simplified tree $\SimplifiedTree = (V_s, E_s)$ rooted at $\TreeRoot$, an ``importance" value is assigned to each branch, and the top $k$ (a user provided threshold) branches are then greedily selected. Here, a branch is a unique path from the root node $\TreeRoot$ to a degree-1 leaf node $l$. It turns out that branch length works well as the importance value. An example on STP data is shown in Fig. \hyperref[fig:5_tracer_injection]{5a}.

\clearpage
\pagebreak
\newpage
\pagestyle{empty}
\beginsupplement
\captionsetup[table]{labelfont={bf},labelformat={default},labelsep=naturebar,name={Supplementary Table}}
\captionsetup[figure]{labelfont={bf},labelformat={default},labelsep=naturebar,name={Supplementary Fig.}}
\section*{\Large Supplementary Methods} \label{sec:Supplementary_information}

\subsection{Implementation in the discrete setting} \label{supp:DM_implementation}
The following provides a more detailed description of how the persistence-guide Morse-based graph framework is implemented.  For even more specifics, please refer to \cite{Forman_DM_1998} for an introduction into Discrete Morse Theory and to \cite{DWW18} for more details on the algorithm we implement.

\begin{algorithm}[h] 
	\caption{$G$ = \DiscreteMorse($\SimpComplex$, $\DensityMap$, $\PersistenceThreshold$)}
		\label{alg:dimorsc}
		\begin{algorithmic}[1]
		\STATE Persistence Computation\\
		- Compute persistence pairings induced by lower-star filtration of $\SimpComplex$ with respect to -$\DensityMap$
		\STATE Obtain Simplified Discrete Gradient Vector Field\\
		- Initialize trivial vector field\\
		- For each persistence pair, perform cancellation if possible and persistence $\leq$ $\PersistenceThreshold$ 
		\STATE Collect Output\\
		- compute the 1-unstable manifold of each critical edge with persistence $>$ $\PersistenceThreshold$
		\RETURN union of 1-unstable manifolds
	\end{algorithmic}
\end{algorithm}

\noindent The Discrete Morse algorithm is traditionally given a triangulation $\SimpComplex$ of the domain and a density function $\DensityMap$ given at the vertices of $\SimpComplex$.  Instead of a triangulation, we take $\SimpComplex$ to be the 2-skeleton(vertices, edges, and squares) of the cubical complex of the domain.  This does not change any part of the algorithm and reduces computation time. Additionally, the user provides the algorithm with a persistence threshold $\PersistenceThreshold$.

\subsection{Step 1}
The first step of the algorithm is to compute the persistence pairings P($\SimpComplex$) by the lower star filtration of $\SimpComplex$ with respect to -$\DensityMap$.  In our implementation, we use DIPHA\cite{DIPHA_Bauer_2014} to compute persistence because it is a distributed persistent homology algorithm that we found minimizes computation time.

\subsection{Step 2}
The second step of the algorithm is to compute the discrete gradient vector field.  As shown in \cite{DWW18}, all that is needed is to calculate the spanning forest that is made up of all negative edges (edges that are paired with a vertex in P($\SimpComplex$)) with persistence less than or equal to $\PersistenceThreshold$. Positive edges (edges paired with a square) and edges with persistence greater than $\PersistenceThreshold$ are not part of the spanning forest. No explicit discrete gradient vector field needs to be computed nor maintained. This step takes linear time once the persistence pairings are computed in Step 1. 

\subsection{Step 3}
The third step of the algorithm is to compute the 1-unstable manifold of each critical edge.  As shown in\cite{DWW18}, for each edge, the 1-unstable manifold is equivalent to the union of the edge with the paths from both vertices to the sink of their corresponding tree in the spanning forest computed in Step 2. The union of all 1-unstable manifolds is outputted by the algorithm.  Please note that the 1-unstable manifold of -$\DensityMap$ is equivalent to the 1-stable manifold of $\DensityMap$.

\subsection{Initial vector-based Morse graph simplification.}
\label{supp:initialvecsimp}
Given the Morse skeleton graph computed in Step 2 of the \myDM{} pipeline, at the beginning of Step 3 of our pipeline, we can perform a flow-vector based graph simplification to remove branches misaligned with underlying flow vectors. 
In particular, we first estimate flow-vectors to get a sense of which direction true neuron branches will flow. To do this, we use a weighted principal component analysis \cite{Vector_simp_Delchambre_2014}. In standard principal component analysis, the principal component represents the direction which explains the most variance of a collection of data points. With weighted principal component analysis, the principal component represents the direction which explains the most variance of the weights of data points. For each vertex in the Morse graph output, we compute this flow-vector for a cubic neighborhood, assigning each point in the neighborhood a weight equal to its intensity.  We then apply Gaussian diffusion to all computed flow-vectors for the sake of smoothing. Note that the flow-vector estimation is only carried out for nodes in the Morse skeleton graph. 

\noindent Next, each path from non-degree 2 to non-degree 2 node in the Morse skeleton graph is computed. Then, for each vertex $v$ in each path $s$, we estimate a path-vector at $v$ w.r.t. $s$ by taking the difference between the two vertices that are 4-hops from $v$ in the path. We compute the cosine between this path-vector and the estimated flow-vector at $v$ and consider this to be the vector score at $v$. The closer the value is to 1, the more the flow-vectors and path-vectors are aligned, meaning the Morse graph skeleton is more aligned with the estimated flows.  On the other hand, the closer the value is to 0, the less the vectors are aligned, indicating the Morse graph skeleton is perhaps significantly deviating from the estimated flows. Note that for each vertex $v$ with degree greater than two, $v$ receives a score for each path it is a part of.

\noindent In addition to these vector scores, a capped intensity value is calculated for each vertex in the Morse skeleton graph. This is simply the minimum of the corresponding voxel value of the vertex and a user-provided value.  For our experiments, the value provided is 1.

\noindent Finally, a score is computed for each path using the vector scores and capped intensity values.  Specifically, for each vertex $x$ in each path $s$, let $c(x)$ return the capped intensity value of $x$, and let $v(x)$ return the vector score of $x$ w.r.t. $s$.  The score of the path is equal to $\int_{s}c(x)(\alpha+v(x))ds$ divided by the length of $s$. $\alpha$ is a user provided weight parameter (a value of zero is used in our experiments).

\noindent Once scores are computed for each path, the simplification process begins.  Only paths below a user-provided threshold are removed. In increasing score order, paths are removed if doing so does not change the number of connected components in the Morse graph skeleton.  If a path is removed, it is possible that paths with lower scores that were not removed will now not alter connectivity if removed, and thus are again tested for removal.  The simplified Morse skeleton graph is then fed into a weighted shortest spanning forest algorithm. 

\subsection{Weighted shortest spanning forest algorithm}
\label{supp:weighted_shortest_spanning_forest}
Given a graph $G = (V, E)$ where each node $v\in V$ is a point in $\mathbb{R}^3$, let $R \subset V$ with $R = \{v_0, ..., v_m\}$ denote the set of roots for spanning forests. 
We then compute a weighted shortest-spanning forest for $G$ as follows: 
For any edge $(u, v) \in E$, its weight is defined as $w(u, v) = \frac{2 d(u, v)}{\DensityMap(u) + \DensityMap(v)}$, where $d(u, v)$ is the Euclidean distance between $u$ and $v$, and $\DensityMap: V \to \mathbb{R}$ is the density map.  
Using these weights (as distances), the algorithm computes the shortest path distances between each root in the root-set $R$ and nodes in $V$. Then for each root $r\in R$, its shortest path tree $T_r$ is spanned by all nodes whose shortest path distance to $r$ is smaller than that to any other node in $R$ (ties are broken arbitrarily). 

\subsection{Score initialization and smoothing}
\label{supp:cal_scores_smoothing}
We now describe how to compute a score $\SimpFinalScore(v)$ for all tree nodes in a given tree $T$ (from the spanning forest), so as to carry out the tree simplification procedure as described in \hyperref[sec:Methods]{Methods} section. 

\noindent Consider a specific tree $T_r$ from the weighted shortest-path forest, with root $r \in R$. 
We first calculate two temporary scores carrying different types of information for each tree node $v$ in $T_r$: one is a density score $\SimpDensityScore(v)$ based on density information, and the other one is a vector score $\SimpVecScore(v)$ based on directional information. The weighted score $\SimpWeightScore$ of each tree node $v$ is the weighted sum of these two temporary scores, i.e., $\SimpWeightScore(v) = \SimpWeight \cdot \SimpDensityScore(v) + (1 - \SimpWeight) \cdot \SimpVecScore(v)$ with $0 \le \SimpWeight \le 1$. Empirically, a large $\alpha$ is used because density is the major indicator for false positives. The details for computing $\SimpDensityScore$ and $\SimpVecScore$ are stated in the following sections. 
After the weighted score $\SimpWeightScore$ is computed for all tree nodes, we further smooth this score to obtain the final score $\SimpFinalScore$. 

\subsection{(1). Calculate density scores for tree/graph nodes.} 
The density score $\SimpDensityScore(v)$ is calculated based on voxel density. For each voxel, we first find its nearest neighbor in the set of tree nodes $V$ and associate it with the nearest neighbor. To avoid associating a voxel to a tree node far away, we set an upper bound $\SimpDensityDistCap$ , thus a voxel will not be associated with any tree node if its distance to the nearest neighbor exceeds this upper bound. 
Empirically, on both injection tracer and fMOST single-neuron datasets, the default upper bound $\SimpDensityDistCap$ is set as the distance between adjacent brain slices (i.e., $1 \upmu$m for the fMOST dataset and $50 \upmu$m for the STP dataset). If a voxel has multiple nearest neighbors, we break the tie arbitrarily. The density score $\SimpDensityScore$ of a tree node $v \in V_s$ is simply the sum of density values of all associated voxels, i.e., 
$$\SimpDensityScore(v) = \sum_{x \in I, d(x, v) = d(x, V_s), d(x, v) \le \SimpDensityDistCap} \DensityMap(x),$$ 
where $I$ denotes the set of all voxels, $\rho$ is the density map and $V_s$ is the node set of the simplified tree (Fig. \hyperref[fig:3_simplification]{3b}).

\subsection{(2). Calculate vector scores for tree nodes.}
Vector scores have already been computed as described in \hyperref[supp:initialvecsimp]{Initial vector-based Morse graph simplification} earlier. 


\subsection{(3). Score smoothing to obtain final score $\SimpFinalScore$.}
Recall that we obtain a combined weighted score $\SimpWeightScore(\cdot)$ from density scores and vector scores. This weighted score will be further smoothed before performing tree simplification to remove local noise. 
In particular, the score of a tree node $v$ will be smoothed by averaging those of its neighbors within $k$-hops (i.e., connected via at most $k$ tree edges to $v$). We further restrict only to neighbors that are ancestors or descendants of $v$ given the root $\TreeRoot$: intuitively, we wish to consider only neighbors of $v$ along the ``same" neural branch. 
\noindent An example with $k = 3$ is shown in Fig. \hyperref[fig:3_simplification]{3c}. The final score of tree node $v$ is $\SimpFinalScore (v) = \frac{1}{|H^k_v|} \sum_{x \in H^k_v} \SimpWeightScore(x)$, where $H^k_v$ denotes the set of $v$'s ancestors and descendants (including $v$ itself) within $k$ hops. In practice, maximum hop $k$ is set to $10$ for both tracer injection and fMOST datasets.

\subsection{Evaluation metrics}
\label{supp:evaluation_metrics}
\subsection{Precision, Recall and F1-score for evaluating fMOST results.}
\begin{figure*}[ht!]
	\centering
	\includegraphics[width=0.8\linewidth]{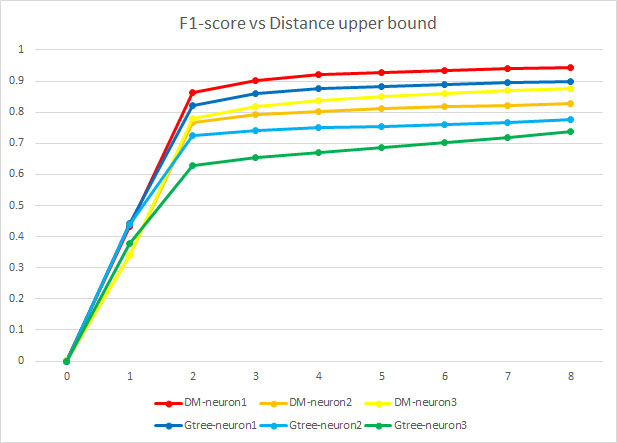}	
	\caption{F1-scores vs different distance bounds. The F1-scores for both \myDM{} and GTree outputs start to stabilize when the upper bound becomes larger than 4 $\upmu$m. Therefore $4 \upmu$m is a reasonable choice for the bound.}
	\label{supp_fig:f1_scores_vs_distance}	
\end{figure*}

For fMOST data, the single neuron reconstruction data were provided as groundtruth. The precision and recall metrics calculated for evaluating fMOST results depend on True Positives (TP), False Positives (FP) and False Negatives (FN). All the outputs were discretized before computing these metrics. We broke any segment with a length greater than 2 pixels so that segment lengths are all roughly 1 pixel. Then, for each node $v$ from the discretized skeletonization, we labelled $v$ as either TP or FP.  $v$ is considered as TP if its nearest neighbor in the human annotation is within $4 \upmu m$. Otherwise, $v$ is a FN. Similarly, a node in the annotation is a FN if there is no predicted node within $4 \upmu m$. Precision and recall are then routinely computed as $Precision = \frac{TP}{TP + FP}$ and $Recall = \frac{TP}{TP + FN}$. The F1-score is the harmonic mean of precision and recall, i.e., $F_1 = \frac{2 Precision \cdot Recall}{Precision + Recall}$. The parameter $4 \upmu m$ was predetermined based on similar reasons as in \cite{Quan_NeuroGPSTree_2016}. The thickness of dendrites near the soma is around $4 \upmu m$ and the curve of F1-score against this parameter (Supplementary Fig. \hyperref[supp_fig:f1_scores_vs_distance]{1}) also supports the choice. The F1-score is stabilized when the parameter reaches $4 \upmu \text{m}$.

\subsection{Region analysis}
\label{supp:regional_analysis}
The brain-wide inter-regional connectivity information was obtained through region analysis on the whole brain tracing data. The 3D volume of the data was registered with the Allen Mouse Brain Atlas at 10 $\upmu \text{m}$ resolution \cite{AllenMouseBrainAtlas}. 
Based on the atlas, the 3D density fields (proofread STP dataset and projected summarizations) were segmented into individual brain regions. Regions were considered "covered" when the total weights of the voxels in these regions were greater than zero. The covered brain regions were ranked based on the total weights of the voxels inside. The two hemispheres were separated and the above described region analysis was performed on each side in addition to the whole-brain analysis. To quantitatively evaluate performance, we calculate a coverage rate for each method on the two hemispheres, and the entire brain. For a given method, the coverage rate is equal to the sum of the weights for regions in which the method's output covers divided by the sum of weights for all regions.


\begin{table*}[ht!]
	\centering
	\begin{tabular}{|c| c c c c c c c |} 
		\hline
		& Method & Precision & Recall & F1 & \makecell{Proofreading \\ time} & \makecell{Total process \\ length (cm)} & \makecell{Normalized proofreading \\ (sec / cm)} \\  
		\hline\hline
		\multirow{6}{*}{Neuron1} 
		& GTree & 0.846&	0.906&	0.875
 & - & \multirow{6}{*}{0.500} & -\\
		& GTree-a1 &0.953&	0.940&	\textbf{\color{red}{0.946}}
  & 23m40s & & 2839.06 \\
		& GTree-a2 & 0.954&	0.929&	0.941
 & 21m40s & & 2599.14 \\
		& DM & 0.900&	0.940&	0.920 & - & &-\\
		& DM-a1 & \textbf{\color{red}{0.957}}	&0.923&	0.940
 & \textbf{\color{red}{15m43s}} & & \textbf{\color{red}{1885.37}} \\
		& DM-a2 & 0.948	& \textbf{\color{red}{0.941}}&	0.944
  & 15m56s & &  1911.37\\
		\hline\hline
		\multirow{6}{*}{Neuron2}	
		& GTree &  0.749&	0.750&	0.750
  & - &\multirow{6}{*}{0.508} & -\\
		& GTree-a1 &  0.818&	0.748&	0.781
  & 20m00s & & 2399.20\\
		& GTree-a2 & 0.807&	0.756&	0.780
 & 21m40s & & 2599.14\\
		& DM & 0.865&	0.749&	0.803
  & - & & -\\
		& DM-a1 &0.866&	\textbf{\color{red}{0.773}}&	0.817&		\textbf{\color{red}{16m00s}} & & \textbf{\color{red}{1919.36}} 
 \\
		& DM-a2 & \textbf{\color{red}{0.944}}	&0.771&	\textbf{\color{red}{0.849}}
& 16m21s & & 1961.35\\
		\hline\hline	
		\multirow{6}{*}{Neuron3} 
		& GTree & 0.621	&0.729&	0.670
  & - & \multirow{6}{*}{0.640} & -\\
		& GTree-a1 & 0.781&	0.744&	0.762
 & 23m21s & & 2801.07\\
		& GTree-a2 &0.726&	0.746&	0.736
 & 22m45s & & 2729.09\\
		& DM & 0.890&	\textbf{\color{red}{0.789}} &	0.836
 & - & &- \\
		& DM-a1 & \textbf{\color{red}{0.939}} &	0.771&	0.847
 & 18m20s &  &2199.27
 \\
		& DM-a2 & 0.925	& \textbf{\color{red}{0.789}}	& \textbf{\color{red}{0.852}}
& \textbf{\color{red}{17m46s}} & & \textbf{\color{red}{2131.29}}\\
		\hline
	\end{tabular}
	\caption{Proofreading is done on GTree and \myDM{} (abbreviated as DM in this table) outputs on all three fMOST neuron regions. We further computed the evaluation metrics on those proofread results with the original ground truth. The suffixes ``a1'' and ``a2'' represents the two annotators for proofreading. The F1-scores are improved after proofreading. The proofreading time spent on the \myDM{} method is also significantly shorter. }
	\label{table:proofreading_fMOST}
\end{table*}

\begin{table*}[ht!]
\centering
\begin{tabular}{|c| c c c c c|} 
\hline
\diagbox[width=10em]{Simplification}{Persistence}&
 128 & 256 & 512 & 768 & 1024 \\ \hline \hline
 0	 & 0.352&	0.368&	0.468&	0.503&	0.445
 \\
 0.05	 & 0.899&	0.911&	0.912&	0.828&	0.639
 \\
 0.10	& 0.909&	0.918&	0.910&	0.825&	0.627
 \\
 0.15	 & 0.910&	0.918	&0.907&	0.818&	0.590
 \\
 0.20	 & 0.912&	\textcolor{red}{\textbf{0.920}} &	0.899&	0.791&	0.584
 \\
 0.25 & 0.902&	0.910&	0.889&	0.788&	0.583
 \\
 0.30	 & 0.897&	0.902&	0.872&	0.776&	0.582
 \\ \hline
\end{tabular}
\caption{This table shows the F1-score of \myDM{} outputs with different simplification and persistence thresholds. The F1-score is calculated based on the region of neuron1. The result of the optimal F1-score uses persistence threshold = 256 and simplification threshold = 0.2. The same thresholds (256, 0.2) are applied to the other two regions.}
\label{table:choose_persistence_simp_threshold}
\end{table*}

\begin{figure*}[ht!]
    \centering
    \includegraphics[width=0.8\linewidth]{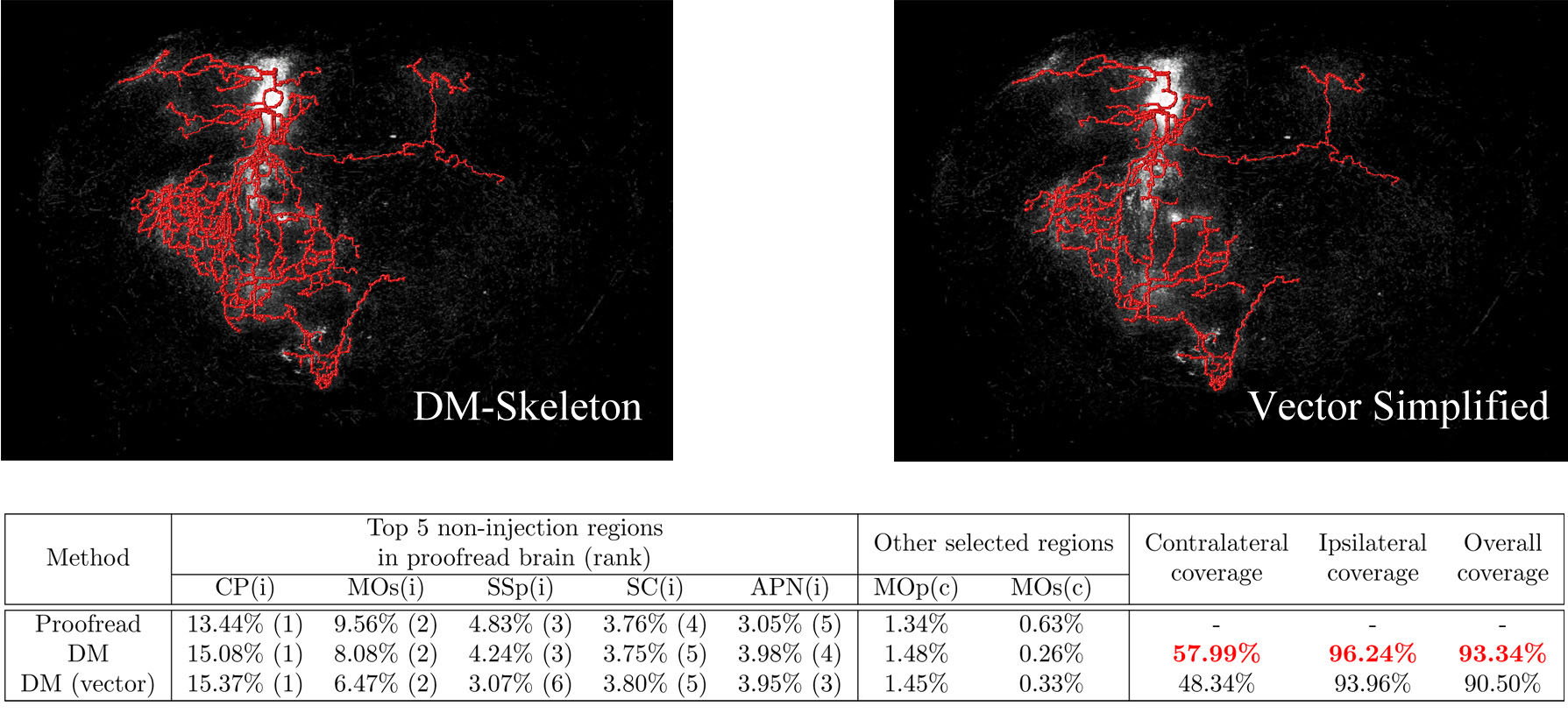}
    \caption{Summarization results of \myDM{} (abbreviated as DM in the table for better presentation) method, and the version after vector simplification. The coverage percentages are shown in the table. We can see that the vector simplification can provide a clearer structure without much loss of coverage. All the selected regions are still covered by the \myDM{} output after vector simplification, but the ranking of regions is slightly distorted.}
    \label{supp_fig:stp_vec_simplification}
\end{figure*}

\end{document}